\documentclass[11pt]{academic_template}
\usepackage[toc,page,header]{appendix}

\usepackage{marvosym}

\usepackage{bbding}

\usepackage{amssymb}
\usepackage{amsmath}

\usepackage{makecell}
\usepackage{algorithm}
\usepackage{algpseudocode}
\usepackage[utf8]{inputenc}
\usepackage[T1]{fontenc}
\usepackage{booktabs}      
\usepackage{array}         
\usepackage{colortbl}      
\usepackage{xcolor}        
\usepackage{caption}       
\usepackage{geometry}     
\usepackage{tabularx,booktabs,colortbl}
\usepackage{graphicx}

\usepackage{listings}
\usepackage{hyperref}
\usepackage{url}
\usepackage{booktabs}
\usepackage{wrapfig}

\usepackage{colortbl}
\usepackage{caption}
\usepackage{subcaption}

\usepackage{amsthm}

 \usepackage{amsmath,amssymb,amsthm,mathtools,bm}
 \usepackage{bbm}

 \theoremstyle{definition}

\setlogoheight{8mm}
\setlogospacing{0mm}
\setsjtublue

\settitlerulethickness{3pt}

\abstractboxon
\setabstractframecolor{gray}
\setabstractbgcolor{gray!10}
\setlogotolineshift{0mm}

\settoprulethickness{2.5pt}
\setbottomrulethickness{1.5pt}

\usepackage{amsmath,amsfonts,bm}

\def\eqref#1{equation~\ref{#1}}
\def\1{\bm{1}}

\DeclareMathAlphabet{\mathsfit}{\encodingdefault}{\sfdefault}{m}{sl}
\SetMathAlphabet{\mathsfit}{bold}{\encodingdefault}{\sfdefault}{bx}{n}

\usepackage{dblfloatfix}
\usepackage{tabularx}

\usepackage{multicol}
\usepackage{multirow}
\usepackage[table]{xcolor}
\usepackage{array}

\newcolumntype{C}[1]{>{\centering\arraybackslash}p{#1}}

\usepackage{natbib}
\setcitestyle{numbers}
\setcitestyle{square}

\usepackage{graphicx}

\usepackage{pifont}
\newcommand{\cmark}{\ding{51}}
\newcommand{\xmark}{\ding{55}}

\newcommand{\teaserfigure}{
\begin{center}
    \includegraphics[width=0.99\textwidth,trim={0.66cm 6.7cm 1.95cm 3.2cm},clip]{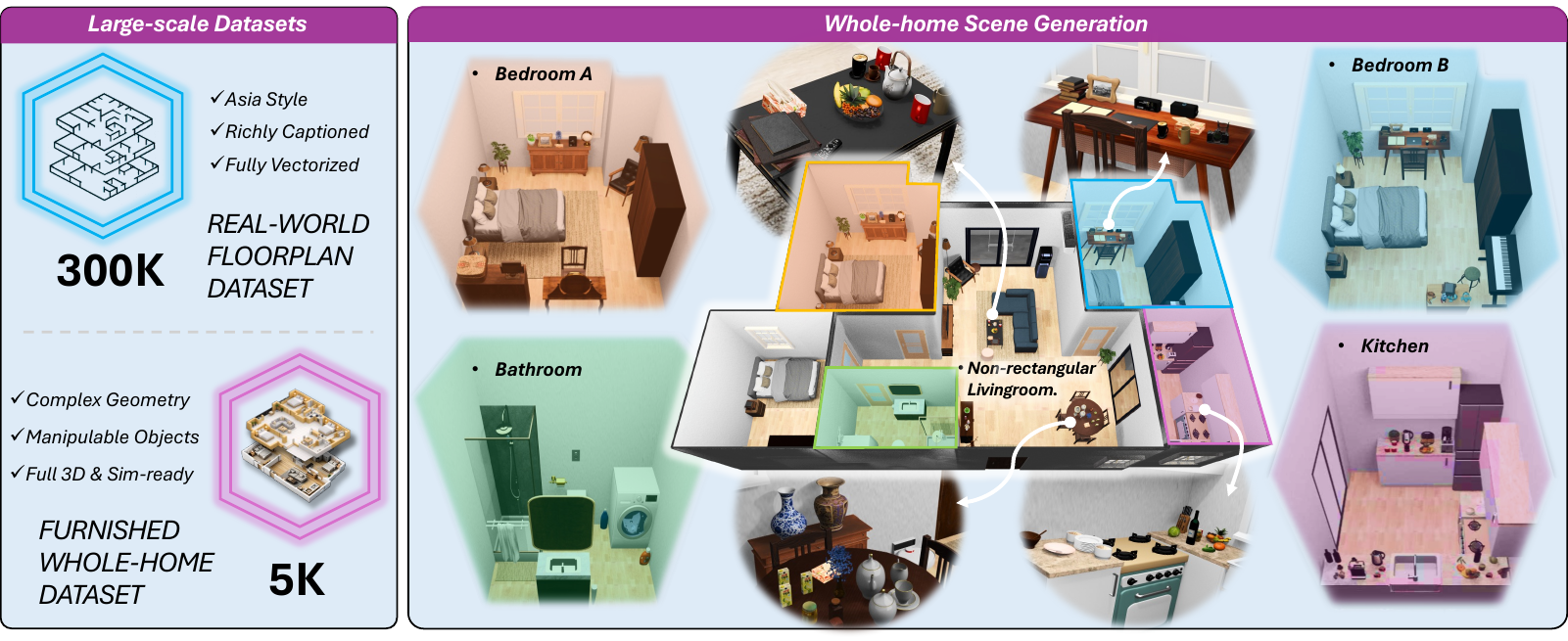}
    \label{fig:teaser}
\end{center}
}

\usepackage{threeparttable}

\title{
\large HomeWorld: A Unified Floorplan-to-Furnished Framework for Generating Controllable, Densely Interactive Whole-Home Scenes
}

\author[1,2,*]{Wenbo Li}
\author[1,2,*,\dagger]{Xiaoliang Ju}
\author[1,2,*]{Zipeng Qin}
\author[2]{Rongyao Fang}
\author[2,1,3]{Hongsheng Li}
\affiliation[1]{Ace Robotics}
\affiliation[2]{CUHK MMLab}
\affiliation[3]{Shenzhen Loop Area Institute}
\contributionfootnote[*]{Equal contribution.}
\contributionfootnote[\dagger]{Project lead.}

\correspondence{Xiaoliang Ju, \email{akira@link.cuhk.edu.hk}; Hongsheng Li, \email{hsli@ee.cuhk.edu.hk}}

\checkdata[Project Page]{\url{https://https://kairos-homeworld.github.io}}

\abstract{
Indoor scene generation is crucial for robot simulation and modern interior design. However, complex layouts together with scarce 3D scene data make learning-based generation challenging. Existing methods often rely on hand-crafted rules or focus on isolated sub-tasks (e.g., floorplan synthesis or single-room furnishing), producing whole-home scenes that lack global coherence, realism, and simulation readiness.
To mitigate these limitations, we propose a unified hierarchical framework that decomposes indoor scene synthesis into controllable stages.
First, we curate a large-scale dataset of 300K real residential floorplans to train a large language model for whole-home floorplan generation.
With detailed descriptions and a K-D tree–based representation, our method enables fine-grained, controllable whole-home floorplan generation.
Building upon the generated whole-home floorplan, we leverage image generation models to draft furniture layouts from multi-level roaming viewpoints, and then generate the layouts of small manipulable objects on different supporting surfaces (e.g. cabinets, desks, and dining tables) for embodied AI simulation. 
During furniture \& object layout generation, a VLM-based refiner iteratively corrects furniture \& object placement, and a 3D generative model enables flexible replacement of individual assets.
We further attach basic physical attributes and simple surface texture and lighting setups to complete the pipeline for embodied AI use.
Experiments and user studies demonstrate that our pipeline produces indoor spaces with greater layout diversity and stronger 3D design appeal, outperforming prior methods on both quantitative and qualitative metrics. Finally, alongside our generation pipeline, we will release the floorplan dataset and 5K fully furnished scenes to the community.}

\begin{document}

\maketitle

\clearpage

\begin{spacing}{0.9}
\tableofcontents
\end{spacing}

\clearpage

\section{Introduction}
\label{sec:intro}

Indoor scene generation is critical for simulation in embodied AI and for modern interior design workflows. As embodied learning and generalist policies advance, agents are expected to navigate, interact, and complete increasingly complex tasks, raising the bar for virtual environments. The target is therefore shifting from isolated, single-room scenes to complete, multi-room homes with globally coherent structure and interactive surroundings. Meeting this demand calls for whole-home scene generation that is not only visually realistic, but also functionally plausible and physically feasible for simulation.

However, scaling to such whole-home generation is fundamentally constrained by data: large-scale, high-fidelity 3D residential scenes are still scarce in public datasets. Existing resources are dominated either by rule-based synthetic datasets~\cite{deitkeProcTHOR️, zheng2020structured3d, savva2017minos}, whose layouts and object distributions are biased by hand-designed priors, or by reconstruction-based datasets~\cite{dai2017scannet, chang2017matterport3d}, which often suffer from incomplete geometry and fragmental surfaces that make reliable layout extraction and asset grounding difficult. As a result, both the quantity and the quality of available 3D data are insufficient to effectively train learning-based generative models for diverse, realistic, and physically feasible multi-room homes.

Existing generative methods struggle under such data regime. Approaches that learn directly in 3D are typically limited by the quantity and fidelity of training scenes, and are computationally expensive, making it difficult to efficiently generate large-scale, whole-home environments~\cite{wu2024blockfusion,ju2024diffindscene}. Meanwhile, rule-based pipelines like~\cite{deitkeProcTHOR️} can guarantee basic validity, but tend to produce repetitive layouts and biased object distributions. Another line of work exploits 2D priors from image generation model to lift room imagery to 3D such as \cite{hollein2023text2room}. However, without a strong 3D prior to anchor the generation, it is difficult to preserve geometric consistency when scaling to whole-home scenes. In practice, the lifted 3D outputs are often incomplete and fragmented, exhibiting  artifacts commonly seen in reconstruction-based 3D data.

To overcome the data bottleneck while meeting the requirements of whole-home, interactive simulation, we propose a unified hierarchical generation framework that combines strong, scalable priors with explicit 3D grounding. Instead of learning whole-home 3D synthesis end-to-end from scarce 3D datasets, we first learn to generate globally coherent 2D floorplan, and then progressively instantiate it into a physically feasible 3D environment with furniture and manipulable objects. A key observation is that even 2D floorplan data—despite being easier to obtain than full 3D scenes—is still relatively scarce in large scale and often lacks clean, standardized representations for learning. We therefore curate and annotate a large dataset of 300K real residential floorplans, and use it to train an LLM for fine-grained, controllable floorplan generation with caption supervision and a K-D tree–based representation.

Conditioned on the generated floorplan, we instantiate the 3D home with a view-roaming, prior-guided synthesis stage. We leverage the diversity and commonsense plausibility of foundation image generation models to propose new objects from roaming camera views, while using 3D constraints derived from the floorplan (e.g., room boundaries, walls, and free space) to anchor the proposals and enforce cross-view 3D consistency. To make the process stable and semantically structured, we adopt a hierarchical roaming policy: we first place large, function-defining furniture under a top-down view (e.g., beds and sofas), and then descend to egocentric views to progressively enrich the scene with smaller items. Finally, to better support embodied AI simulation, we distribute manipulable objects throughout the home. Throughout this process, an VLM-based recursive refiner corrects implausible placements and constraint violations, and a 3D generative module enables flexible asset replacement to increase object diversity and rendering variation without breaking scene coherence.

In summary, our contributions are two-fold:
\begin{itemize}
\item \textbf{A whole-home indoor scene generation pipeline.} We propose a unified hierarchical pipeline for controllable whole-home generation under limited 3D data. It (i) trains LLMs to generate precise and controllable floorplans via a K-D tree representation and detailed captions; (ii) leverages foundation image models as strong 2D priors to propose objects through hierarchical view roaming while maintaining 3D consistency and physical adherence; (iii) employs an VLM-based recursive refiner to correct construction errors; and (iv) supports flexible 3D asset replacement via a 3D generative module to increase object diversity and rendering variation.
\item \textbf{A large-scale whole-home indoor scene dataset.} We curate and will publicly release a dataset of \textbf{300K} annotated real-world residential floorplans, together with \textbf{5K} fully furnished, high-quality whole-home 3D scene samples, providing a new dataset that directly mitigate the data scarcity bottleneck in whole-home layout generation. We plan to continuously expand this dataset in subsequent releases.
\end{itemize}

\section{Related Work}

\begin{table*}[ht]
\caption{{\textbf {Indoor Scene Resources Comparison.}} \textbf{Rec.} denotes reconstruction-based real-world datasets. \textbf{S.}/\textbf{H.} denote the number of furnished \textbf{Scenes} (individual, often room-level areas) and \textbf{Homes} (unified, with multiple rooms).\textbf{Sim-ready} denotes whether a dataset provides fully 3D scenes that are directly instantiable in a simulator/rendering engine and supports object-level manipulation. \textbf{MObj.} reports manipulable objects per scene on average when available. \textbf{n/r} means not reported and ``--'' means not applicable. \textbf{Collection} indicates datasets aggregated from multiple sources.}
\centering
\footnotesize
\setlength{\tabcolsep}{4pt}
\renewcommand{\arraystretch}{1.15}

\begin{threeparttable}
\resizebox{\textwidth}{!}{%
\begin{tabular}{lcccccccc}
\toprule
\multicolumn{8}{c}{\bf PART A: STATIC / RELEASED DATASET}\\
\midrule
\multirow{2}{*}{Resource} &
\multirow{2}{*}{Source / Training Data} &
\multirow{2}{*}{Scope} &
\multirow{2}{*}{Sim-ready} &
\multirow{2}{*}{Curation} &
\multicolumn{3}{c}{Data Scale / Scalability} \\
\cmidrule(lr){6-8}
 &  &  &  &  & Floorplan & Furnished S./H.  & MObj. \\
\midrule
LI-FULL~\cite{LIFULLNII2017LIFULLHRFP}         & Real & Home & \xmark & Raw & 5M & - & - \\
RPLAN~\cite{Wu_DeepLayout_2019}           & Real & Home & \xmark & Manual & 80K & - & - \\
MSD~\cite{van2024msd}   & Real & Home & \xmark  & Manual  & 5.3K & - & - \\
ResPlan~\cite{abouagour2025resplan}        &Real  & Home & \xmark & Manual  & 17K & - & - \\
ScanNet~\cite{dai2017scannet}    & Real(Rec.) & Partial  & \xmark & Manual & - & 1.5K S. & n/r \\
MatterPort3D~\cite{chang2017matterport3d}    & Real(Rec.) & Home & \xmark & Manual & 90 &  90 H. & n/r \\
3D-FRONT~\cite{fu20213d}        & Designed & Home  & \cmark & Manual & 6.8K & 6.8K H. & n/r \\
Structured3D~\cite{zheng2020structured3d}    & Designed & Home & \xmark & Programmatic & 3.5K & 21K S. & n/r \\
InternScenes~\cite{zhong2025internscenes}    & Collection  & Mixed & Partial & Hybrid & - & 48K S. & 8  \\
SceneVerse~\cite{jia2024sceneverse}      & Collection  & Mixed & Partial  & Hybrid & - & 68K S. & n/r \\
ProcTHOR-10K~\cite{deitkeProcTHOR️}      & Synthetic(Rules) & Home & \cmark & Programmatic & 10K & 10K H. & n/r\tnote{*} \\
\textbf{Ours (Released)}   & FP.(Real) + Obj(Gen.) & Home & \cmark & Hybrid & 300K & 5K H. & >15 \\
\midrule
\multicolumn{8}{c}{\bf PART B: GENERATION PIPELINE}\\
\midrule
Floorplan-LLaMA~\cite{yin2025floorplan} & Collection & Home & \xmark & LLM  & \cmark & \xmark & \xmark \\
Floorplan-Diffusion~\cite{xu2025floorplan} & Real & Home & \xmark & Diffusion & \cmark & \xmark & \xmark \\
LayoutGPT~\cite{feng2023layoutgpt} & None & Partial & Partial & LLM & \xmark & \cmark  & \xmark \\
Holodeck~\cite{yang2024holodeck} & Collection & Home & $\checkmark$ & LLM & \cmark & \cmark   & \xmark \\
LayoutVLM~\cite{sun2025layoutvlm} & Designed & Partial & Partial & VLM & \xmark & \cmark & \xmark \\
ProcTHOR~\cite{deitkeProcTHOR️}  & Synthetic(Rules) & Home & $\checkmark$ & Programmatic & \cmark & \cmark   & \xmark \\
Infinigen~\cite{raistrick2024infinigen}  & Synthetic(Rules) & Home & $\checkmark$ & Programmatic & \cmark & \cmark & \xmark \\
PhyScene~\cite{yang2024physcene} & Designed & Home & $\checkmark$ & Diffusion & \cmark & \cmark & \xmark \\
ChOrD~\cite{su2025chord} & Designed & Home & $\checkmark$ & Diffusion & \cmark & \cmark & \xmark \\
EmbodiedGen~\cite{wang2025embodiedgen} & Collection & Mixed & $\checkmark$ & Diffusion & \xmark & \cmark & \xmark \\
\textbf{Ours (Pipeline)}   & Real floorplan only & Home & \cmark & MLLM & \cmark & \cmark & \cmark \\

\bottomrule
\end{tabular}%
}
% \begin{minipage}{\linewidth}
\begin{tablenotes}
\scriptsize
% % \footnotesize
\item[*] ProcTHOR-10K does not report manipulable-object counts per scene; it provides an asset library containing 1633 \\interactive instances.
\end{tablenotes}

\label{tab:dataset_compare}
% \vspace{-2em}

\end{threeparttable}
\end{table*}

\noindent\textbf{Indoor scene dataset.} We summarize representative datasets in ~\cref{tab:dataset_compare} and group them by data source, scope,  simulator readiness, and curation, together with key scale statistics.
Early real-world floorplan dataset such as LI-FULL~\cite{LIFULLNII2017LIFULLHRFP}, RPLAN~\cite{Wu_DeepLayout_2019}, MSD~\cite{van2024msd}, and ResPlan~\cite{abouagour2025resplan} provide large-scale residential layouts, but they are typically limited to 2D floorplans and therefore are not directly usable for embodied simulation. 
In contrast, reconstruction-based datasets (e.g., ScanNet~\cite{dai2017scannet} and Matterport3D~\cite{chang2017matterport3d}) offer real 3D scans with semantic annotations; however, these scans often cover partial spaces and generally do not provide instance-separated, manipulable assets required by interactive simulators, making them less ``sim-ready'' in our definition.

To bridge the gap, several designed or synthetic datasets provide furnished 3D indoor scenes. 3D-FRONT~\cite{fu20213d} contains thousands of curated residential scenes with explicit object instances and renderable assets, and is commonly used for 3D scene synthesis and rendering. Structured3D~\cite{zheng2020structured3d} programmatically generates large-scale indoor scenes with ground-truth structure, but it is not primarily designed for interactive manipulation. 
More recent ``collection'' datasets, including InternScenes~\cite{zhong2025internscenes} and SceneVerse~\cite{jia2024sceneverse}, aggregate heterogeneous sources to scale up scene diversity; while they provide many furnished scenes, simulator readiness and manipulability vary across subsets (hence marked as partial). 
ProcTHOR-10K~\cite{deitkeProcTHOR️} procedurally generates fully 3D houses that are directly usable in simulation; although it does not report per-scene manipulable-object counts, it provides an interactable asset library of 1633 instances. 

Overall, existing datasets trade off between scale, realism, and direct usability for embodied simulation. Our dataset targets this gap by combining large-scale real floorplans with generated object assets to produce sim-ready, furnished 3D houses with hybrid curation.

\noindent\textbf{3D indoor scene generation.} Prior work on 3D indoor scene generation can be broadly grouped into the following categories. 1) Procedural rule-based generation constructs scenes by executing pre-defined grammars, priors, or handcrafted constraints (e.g., \cite{deitkeProcTHOR️,raistrick2024infinigen}), which can yield controllable outputs but is often bottlenecked by a finite rule set and asset library—resulting in limited diversity.  
2) Direct 3D-domain generation like \cite{ju2024diffindscene,bokhovkin2025scenefactor,wu2024blockfusion,huang2025midi} naturally satisfies geometric constraints, but scaling to whole-home scenes remains challenging due to computational cost and the limited availability of diverse 3D training data.
3) Compositional generation that first predicts a layout and then fills it with 3D assets has therefore become a common paradigm. \cite{yang2024holodeck,ccelen2024design,liu2025worldcraft} use LLMs to generate whole-home floorplans and asset layouts, but remains at a relatively preliminary stage. Since native LLMs have limited spatial understanding, the resulting layouts are also limited in terms of plausibility and diversity.
More methods focus on room-level scene, such as \cite{tang2024diffuscene, feng2023layoutgpt, sun2025layoutvlm, bucher2025respace, yang2024llplace}. While these methods work well for composing a full home from individually generated rooms, indoor environments exhibit strong global coherence (e.g., functional adjacencies and consistent style/scale). As a result, approaches limited to local context often struggle to maintain whole-home consistency.
4) Another promising direction is “roaming” or incremental whole-scene creation via 2D-lifting approaches such as \cite{hollein2023text2room, chung2023luciddreamer, chen2025housecrafter}; however, because these methods often lack strong 3D constraints, they tend to suffer from poor geometric consistency and are hard to deploy in simulation.

Our pipeline adopts a compositional approach that integrates roaming-based techniques while mitigating their limitations. We enforce explicit 3D constraints and employ a reflective loop for iterative correction, ensuring strong 3D consistency throughout generation. 
In addition, most prior work does not explicitly target simulation-ready environments with plentiful manipulable objects for embodied AI.
In contrast, we generate complete homes hierarchically, progressing from a floorplan, to a furniture layout, and finally to manipulable small objects, producing coherent, fully populated, simulation-ready scenes.

\begin{figure}[t]
    \centering
    \includegraphics[width=\textwidth, trim={2.4cm  9cm 0.7cm 4.5cm}, clip]{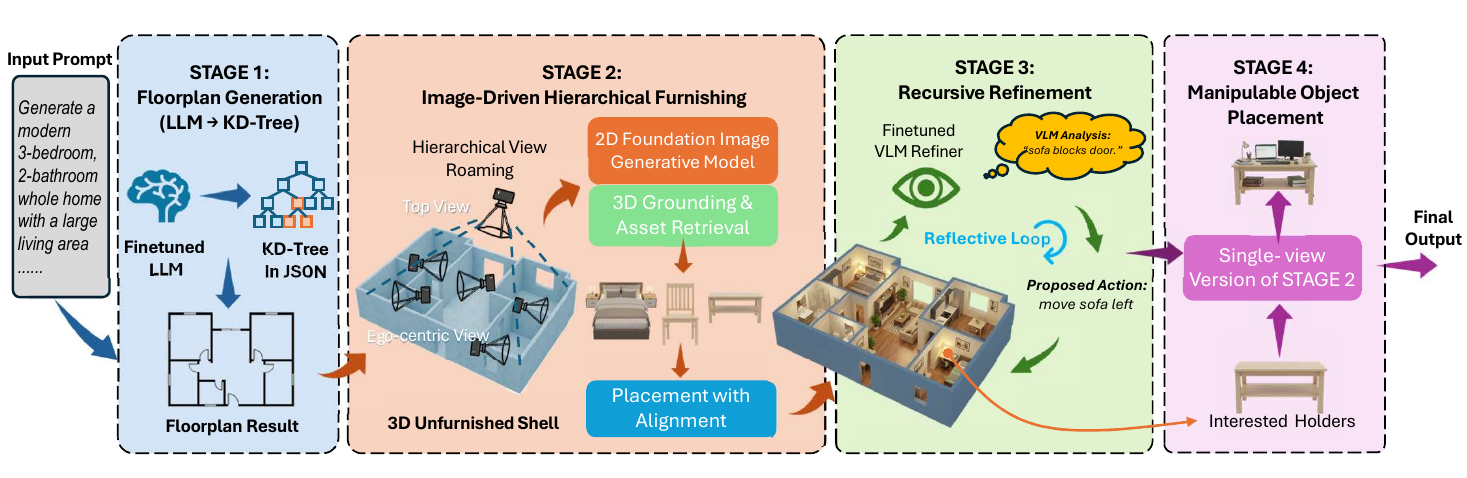}
    \caption{{\bf Overview of the whole-home indoor scene generation pipeline.} Our system turns a text prompt into a complete indoor scene in four stages: 1) build a large-scale floorplan dataset and train a prompt-conditioned generator; 2) given the floorplan, create an unfurnished 3D shell and progressively place furniture via hierarchical roaming with grounding/retrieval; 3)  improve the scene using a finetuned VLM-based refiner to fix violations in a reflective loop; 4) add manipulable objects in feasible surface tops (e.g. dining tables, desks, cabinets, etc.).}
    \label{fig:pipeline}

    \end{figure}

\section{Methodology}

As illustrated in \cref{fig:pipeline}, our system enables users to describe a desired home layout with a natural-language prompt and automatically synthesizes a complete indoor scene. The workflow can be divided into four stages.
First, we describe how we build a real-world large-scale floorplan dataset and train a prompt-conditioned floorplan generator. 
Second, conditioned on the floorplan, we introduce a hierarchical roaming strategy to populate the scene progressively. Before starting, we first build 3D unfurnished shell as the explicit 3D constraint for the roaming. 
Then the system prioritizes the placement of large, room-defining furniture, and then adds more objects, using grounding and asset retrieval to ensure geometric and semantic consistency.
Third, we employ a recursive refinement loop based on a finetuned VLM, which inspects the rendered scene to detect violations (e.g., collisions or blocked pathways), and then acts as a refiner agent that proposes and executes corrective actions. Finally,
we use the similar strategy as Stage 2 to add manipulable objects in user-specified areas of interest.

\subsection{Floorplan Dataset Curation and LLM-based Floorplan Generation}
Current floorplan datasets are limited in scale, complexity, diversity, and realism~\cite{Wu_DeepLayout_2019,van2024msd,abouagour2025resplan}. Yet for applications such as robot navigation, a large-scale and high-fidelity dataset is essential to enable robust generalization to real-world environments. In this section, we develop an automatic procedure for extracting vectorized data from carefully-curated large-scale real-world floorplan images from professional architectural portfolios. We also devise a K-D tree representation for effective training of floorplan generation model.

\begin{figure}[htbp]

    \centering
    \includegraphics[width=\textwidth, trim={1cm, 11cm, 5cm, 4cm},clip]{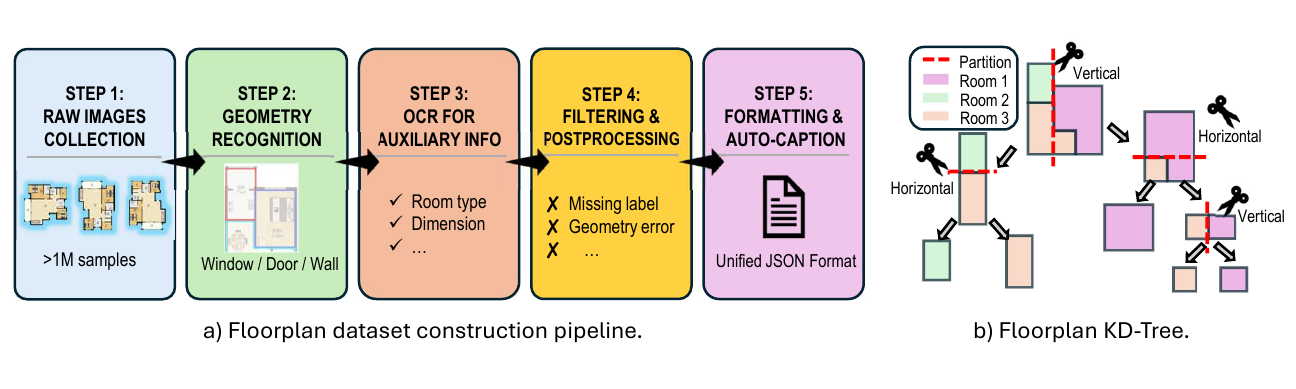}

    \caption{\textbf{(a)} Floorplan dataset construction pipeline and \textbf{(b)} the KD-tree representation for floorplan used in our generator.}
    \label{fig:l1_procedure}

\end{figure}

\noindent\textbf{Floorplan dataset curation.}
We construct a large-scale floorplan dataset in five steps (\cref{fig:l1_procedure}(a)), which transforms raw floorplan images into structured representations with three key attributes: (1) the positions of architectural elements, such as windows and doors; (2) room geometries and spatial layouts; and (3) semantic room-type labels.

In step 1, we collect 1.084 million floorplan images from an online real-estate repository. These images contain common architectural elements, including doors, windows, walls, room labels, and dimension annotations. In step 2, we train a detection model on 2K manually annotated images to identify doors, windows, walls, and dimension chains. The detected bounding boxes are converted into centerlines, and nearby endpoints are merged to reconstruct the vectorized floorplan structure.

In step 3 and 4, we apply OCR to extract room labels and dimensional annotations, followed by quality filtering to discard noisy or unreliable samples. This yields validated floorplans with reliable geometric and semantic information.

In step 5, we further derive structural annotations such as room connectivity from shared doors and adjacent walls, and assign unique identifiers to all entities, including rooms, doors, windows, and walls. Based on the resulting geometric, semantic, and topological information, we automatically generate detailed textual captions for each floorplan.

These captions are used to construct training data for the LLM-based generator. Specifically, we randomly sample and combine caption components, such as room counts, boundary information, positional constraints, connectivity relations, and attachment requirements, to form diverse natural-language prompts, while the corresponding floorplan representation serves as the target output. This procedure allows each floorplan to generate multiple prompt--target pairs with varying levels of difficulty. The final dataset contains 314K validated floorplans, all fully vectorized and paired with detailed structural captions.

\noindent\textbf{LLM-based floorplan generator with K-D tree representation.}
We formulate floorplan generation as a text-to-structure prediction task. The input to the LLM is a natural-language description of the design requirements, including the apartment boundary, room program, positional constraints, adjacency relations, and placement requirements for openings or attachments. The output is not direct geometric primitives such as room polygons; instead, the model predicts a structured \texttt{json} file that explicitly encodes the floorplan as a binary K-D tree, together with attachment information.

In this representation, the interior space is recursively partitioned into axis-aligned subregions by alternating vertical and horizontal cuts (Fig.~\ref{fig:l1_procedure}(b)). Each internal node stores the split axis and split coordinate, while each leaf node corresponds to a spatial cell and is assigned semantic information such as room identity and room type. Attachments, including doors, windows, and related elements, are represented separately in the same output \texttt{json}. Because the entire layout is serialized as a hierarchical text structure, it is naturally compatible with the autoregressive generation process of LLMs.

Compared with directly predicting polygon coordinates and openings, this K-D tree-based output space is much more structured and easier to constrain. In particular, it avoids many common geometric errors, such as polygon overlap, and produces layouts that can be deterministically converted back into conventional floorplan representations for downstream use. Since generic pretrained LLMs do not possess sufficient domain knowledge for this task, we finetune the model on our floorplan dataset. Ablation results that validate the effectiveness of this representation are provided in \cref{sec:ablation}.

\subsection{Image-driven Hierarchical Furnishing}
We adopt a 2D-lifting approach realized through hierarchical indoor roaming, primarily because existing 3D furniture-layout datasets are insufficient in both quantity and quality as aforementioned. We leverage the strong semantic priors of 2D foundation generative models to enable open-vocabulary object placement with improved semantic plausibility. Our system then grounds the generated results into precise 3D bounding boxes, parameterized by semantic label, physical dimension and pose. The roaming consists of two stages: \textit{top-down roaming} initializes coarse object placement at the scene level, while \textit{ego-centric roaming} focuses on local regions to refine object geometry.

\noindent\textbf{Unfurnished shell as 3D constraints for roaming.} Before the roaming starts, we first instantiate an empty, unfurnished shell in Blender with basic lighting and material settings. This shell is used to impose explicit 3D constraints during roaming, which is especially important for irregular, non-rectangular room layouts as shown in the teaser image. With the consistent geometry, it reduces perspective-induced artifacts and 3D inconsistencies caused by fully free roaming (e.g., drifting geometry in  Text2Room~\cite{hollein2023text2room}). In addition, the lighting-driven shading provides a natural prior for the subsequent 2D generation stage, leading to more coherent appearance and smoother view-to-view transitions.

\noindent\textbf{Global layout generation via top-down view.} 
The key idea is to exploit a structure-aware top-down view as an intermediate representation, since major furniture placement is largely governed by global constraints: room boundary, doors, windows, and inter-room connectivity. 
Given the empty shell, we render a top-down view image $I_{\text{top}}$. We then annotate $I_{\text{top}}$ with doors and windows, and for each door we also mark the adjacent room, because these cues strongly influence the furniture layout. 
% \textit{Conditional synthesis.}
Conditioned on these structural priors, we apply an image inpainting model to $I_{\text{top}}$ to synthesize a furnished top view $\hat{I}_{\text{top}}$, leveraging 2D foundation priors for more realistic, prompt-controllable furniture arrangements.
% \textit{Parsing and lifting.}
After a lightweight validation step to discard infeasible generations, we use a VLM for open-vocabulary category recognition and SAM-3~\cite{carion2025sam} for instance masks and 2D boxes. This initial layout of major furniture will be further refined in the following stage.

\noindent\textbf{Detail refinement via ego-centric view.} 
While the global top-down view provides a reliable initialization for major furniture, it is inherently limited by occlusion and scale ambiguity, making it difficult to recover small objects and wall-attached items (e.g., bins, hooks, racks, and upper/lower cabinets in kitchens). We therefore introduce an ego-centric roaming stage to enrich the scene with fine-grained details.

\textit{Viewpoint selection.} To cover the entire room with limited renders, we compute a compact set of viewpoints using a heatmap-based viewpoint selection strategy. Concretely, we discretize the floor plane into a grid and sample candidate camera poses along the room boundary and interior, subject to a fixed field-of-view and a safe distance to walls. Each candidate view contributes visibility to a subset of grid cells; we accumulate these contributions into a coverage heatmap. We then greedily select viewpoints by repeatedly choosing the view that covers the currently least-observed regions (equivalently, maximally reduces the uncovered heat), updating the heatmap after each selection.

\textit{Ego-centric synthesis and 3D grounding.} For each viewpoint $k$, we render an ego-centric image $I_k$ and apply inpainting to synthesize additional secondary objects, producing $\hat{I}_k$. We then use SAM-3~\cite{carion2025sam} to extract instance masks for newly introduced objects and employ SAM-3D~\cite{chen2025sam} to reconstruct their 3D geometry. 
We apply a geometric alignment module to integrate these newly reconstructed 3D assets into the existing room environment. By moving the viewpoint, this iterative ego-centric roaming process provides 360-degree coverage, resulting in a densely populated and highly realistic 3D indoor environment.

\subsection{Recursive layout refinement}

Since both 2D synthesis and 3D grounding are inherently error-prone, the generated layouts may contain physical, semantic, and structural violations, such as blocked doorways, boundary-crossing objects, object collisions, and floating objects. To address these issues, we introduce a lightweight VLM-based refinement module that performs closed-loop verification and correction for each candidate room layout, thereby improving the robustness of the proposed generation pipeline.  We begin with the iterative refinement formulation, since it defines the sequential correction task that the refiner is trained to solve, and then describe the construction of the corresponding fine-tuning dataset.

\noindent\textbf{Iterative refinement formulation.} For each candidate room layout, we render a top-view image and pair it with the corresponding structured 3D layout description, which encodes object categories, positions, sizes, rotations, and room-level architectural elements. These multimodal observations are provided to the VLM refiner, which evaluates the layout under physical plausibility, semantic consistency, and structural constraints. Rather than treating refinement as a one-shot validation problem, we formulate it as a sequential decision-making process: at each iteration, the refiner observes the current layout state and predicts a structured corrective action that identifies the target object and specifies the required edit, such as translation, rotation, or a combined transformation. The predicted action is then deterministically applied to the 3D bounding-box layout representation, after which the updated layout is re-rendered and re-checked. This iterative loop continues until the layout passes validation or a predefined iteration limit is reached.

\noindent\textbf{Fine-tuning data construction.} Following the above sequential refinement formulation, we construct a dedicated supervision corpus to train the VLM refiner for reliable multi-step layout correction. The dataset is designed to expose the model to invalid intermediate layouts and teach it to predict the next corrective action that best improves the scene. It is built from three complementary components: \textit{corrupted layout construction}, \textit{oracle-labeled repair actions}, and \textit{model-in-the-loop samples}.
\begin{itemize}

\item \textit{Corrupted Layout Construction}. Starting from clean room layouts that satisfy basic physical, semantic, and structural constraints, we inject controlled perturbations to synthesize corrupted states. These perturbations cover common failure modes observed in the generation pipeline, including boundary violations, object collisions, doorway blocking, implausible rotations, scale inconsistencies, and category-level errors. Since each perturbation step is explicitly recorded, the resulting clean-to-corrupted trajectories naturally provide reverse supervision for iterative layout repair.

\item \textit{Oracle-labeled Repair Actions}. For each corrupted state, we further generate high-quality action labels using an oracle action generation pipeline. The oracle first proposes a set of candidate corrections, such as translation, rotation, combined transformation, and displacement along the minimum separation direction. Each candidate is then evaluated by a verifier according to residual errors, newly introduced violations, movement cost, and semantic plausibility. The highest-scoring candidate is selected as the ground-truth next action for the current layout state.

\item \textit{Model-in-the-loop Samples}. To improve robustness under closed-loop inference, we additionally collect model-in-the-loop samples. After training an initial refiner, we roll it out on corrupted layouts and collect the intermediate states produced by the model during iterative correction. These states are relabeled by the oracle and added back to the training set, allowing the model to learn from its own failure cases and reducing the distribution gap between supervised training and deployment. Each training sample consists of a top-view rendering, the corresponding structured 3D layout description, the current error state, optional action history, and the oracle-labeled correction action. This dataset construction enables the refiner to learn stable multi-step correction behavior under physical, semantic, and structural constraints.
\end{itemize}

\subsection{Manipulable Object Placement}

Although the top-down and ego-centric roaming stages recover coherent layouts for large furniture, many supporting surfaces, such as tables, desks, countertops, shelves, and nightstands, may still remain unnaturally empty. This reduces both visual realism and functional richness, since small manipulable objects provide not only clutter cues for everyday environments but also potential interaction targets for embodied tasks. To address this issue, we introduce a surface-centric object placement strategy that populates supporting furniture with plausible small objects.

\noindent\textbf{Surface Object Synthesis.} Given a generated room, we first identify furniture items that can support small objects according to their semantic categories and geometric properties, such as the existence of a horizontal surface and sufficient available area. For each selected furniture item, we construct a local surface canvas conditioned on the room type, furniture category, and the target support region. An inpainting model is then used to synthesize realistic surface-level object arrangements by leveraging learned priors on object co-occurrence and spatial organization. For example, a desk may be populated with books, a laptop, a lamp, and stationery, while a kitchen countertop may contain bowls, bottles, utensils, or small appliances.

\noindent\textbf{Physical Attribute Assignment.} To make synthesized manipulable objects physically meaningful and simulation-ready, we use PhysX-Anything~\cite{cao2025physx} to infer object-level and part-level physical attributes. For each object, PhysX-Anything~\cite{cao2025physx} predicts its semantic category, dimensions, part decomposition, and material properties, including density, Young’s modulus, and Poisson’s ratio. In our pipeline, the predicted density and mesh volume are used to estimate approximate object mass, while part-level material properties support the assignment of component-wise simulation parameters. These attributes are further used during layout recovery and validation to assess support validity, object stability, collision consistency, and plausible interaction behavior, improving the physical grounding of generated small-object arrangements for downstream simulation and embodied interaction.

\noindent\textbf{3D Layout Recovery and Filtering.} The synthesized surface image is subsequently parsed by a VLM into a set of manipulable objects, including their semantic categories, approximate scales, relative positions, and orientations. These predictions are first represented in the local coordinate system of the supporting furniture, and are then transformed into the global room coordinate system according to the furniture pose and physical dimensions. This formulation ensures that small objects remain spatially aligned with their host furniture even when the furniture is rotated or placed at different locations in the room.

Before inserting these objects into the 3D scene, we perform lightweight physical filtering to improve plausibility. Specifically, we remove or adjust objects that exceed the support boundary, produce severe inter-object collisions, float above the supporting surface, or violate category-level support constraints. We also preserve the explicit support relation between each small object and its host furniture, which facilitates downstream scene graph construction, task generation, and embodied interaction. As a result, the generated rooms are enriched with realistic and functionally meaningful object-level details while maintaining basic physical consistency.

\subsection{Shell Texture Generation and Basic Lighting}
\noindent\textbf{Surface Texturing.} For the indoor shell, including walls, floors, and ceilings, textures are obtained from two sources. When suitable assets are available in the 3D-FRONT~\cite{fu20213d} repository, we directly reuse those materials. Otherwise, the textures are synthesized using a 2D image generation model and then applied to the corresponding surfaces. Since this component is not a core contribution of our work, we adopt this simple strategy to enrich the appearance diversity of indoor shells without introducing a dedicated material generation pipeline.

\noindent\textbf{Lighting Setup.} For indoor lighting, we adopt rule-based generation methods. A primary ceiling light is placed near the center of each room to provide the main illumination, while auxiliary and decorative lights are configured using predefined heuristics based on the room layout. These rules determine the placement, color temperature, and relative brightness of secondary light sources. The final renderable parameters are assigned from predefined templates to ensure stable and consistent rendering.

\section{Experiments}
\begin{figure}[h!]
    \centering

    \includegraphics[width=0.97\textwidth,trim={0cm 0cm 1.3cm 0cm},clip]{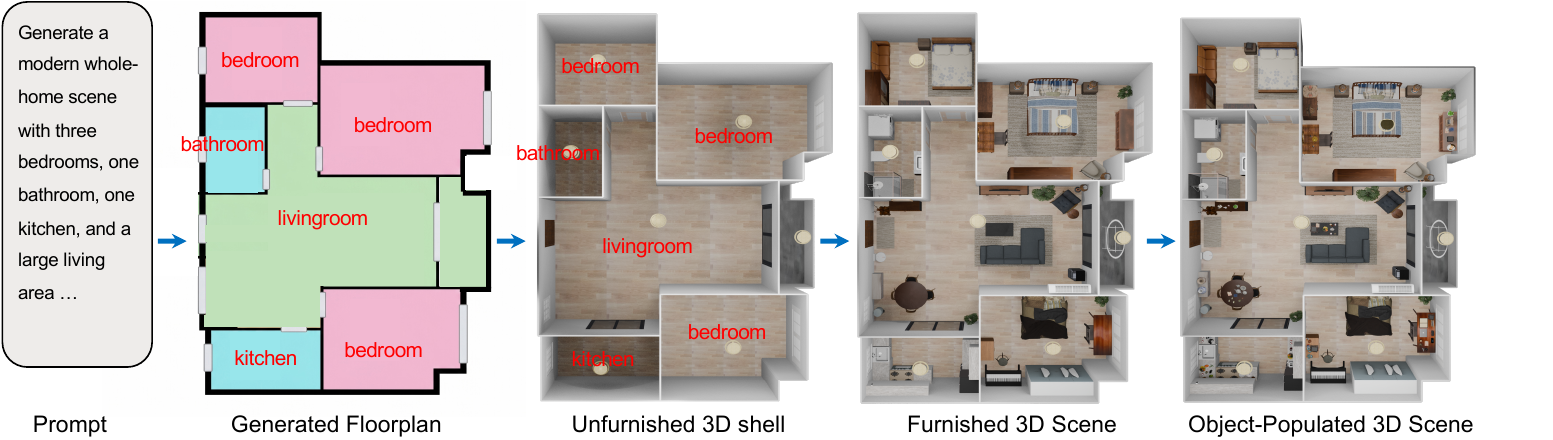}
    \caption{Overview of the whole-home scene generation.}
    \label{fig:stagewise_generation}

\end{figure}
\Cref{fig:stagewise_generation} provides an intuitive overview of our whole-home scene generation pipeline. Given a natural-language prompt, our method progressively generates a floorplan, instantiates an unfurnished 3D shell, produces a furnished 3D scene, and finally enriches the environment with manipulable objects. Following this pipeline, our experiments are organized into three parts. First, we evaluate floorplan generation to demonstrate the effectiveness of our proposed K-D tree-based representation on our dataset. Second, we report results on floorplan-conditioned whole-home generation. Third, we conduct a comprehensive ablation study. Both quantitative and qualitative evaluations show that our method outperforms existing approaches.

\subsection{Floor Plan Generation}

\noindent \textbf{Experiment setup.} We split our dataset into 281K training samples, 15K validation samples, and 15K test samples. We finetune Qwen3-4B-Instruct~\cite{qwen3} using supervised finetuning with LoRA~\cite{hu2022lora} on our training corpus.

We formulate floorplan synthesis as a constrained instruction-following task. The prompt specifies following factors of the target floorplan: a) dimensions; b) room count and types; c) door/window placement; d) a structural output format requiring a K-D tree with axis-aligned (horizontal or vertical) partitions; e)~spatial constraints regarding room adjacencies and connections.

\noindent \textbf{Evaluation metrics.} We introduce two quantitative metrics to assess the generated floorplans: 1) {\it graph validity}, the fraction of generated floorplans whose connectivity graph is fully connected, measuring basic topological navigability; 2) {\it graph diversity}: the number of distinct connectivity graph structures generated for a fixed room count, measuring topological variety. 
We report typed diversity (room types included) and untyped diversity (topology only).

\begin{table}[htbp]

\caption{
    \textbf{Quantitative comparison of floorplan generation.} Our model generates more accurate and diverse room  topologies than the selected baseline.
}
\vspace{-5pt}
\centering
\scriptsize
\begin{tabular}{lccc}
\toprule
             & \multicolumn{1}{c}{\multirow{2}{*}{\hspace{25pt}Graph Validity\hspace{25pt}}} & \multicolumn{2}{c}{Graph Diversity} \\  
\cmidrule{3-4} 
             & \multicolumn{1}{c}{}                            & Typed         & Untyped        \\ 
\midrule
Procthor-10K~\cite{deitkeProcTHOR️} & 72.6\% & 6.2             & 2.3              \\ 
\midrule
Ours         & \bf{87.7\%} & \bf{22.4}             & \bf{6.9}              \\
\bottomrule
\end{tabular}

\label{tab:fp_gen_com}
\end{table}

\noindent \textbf{Quantitative results.} We randomly sample 500 floorplans from our finetuned Qwen3 model and compare them against generated samples from Procthor-10K\cite{deitkeProcTHOR️}. For graph validity, our model achieves a 15.1\% improvement over Procthor-10K (see~\cref{tab:fp_gen_com} column 1), demonstrating its ability to learn accurate opening placements (e.g., doors) that yield fully connected floorplan topology. For both typed and untyped graph diversity, we compute the numbers of distinct connectivity graphs for samples containing one to ten rooms - a range representative of typical residential layouts - and report the average. As shown in \cref{tab:fp_gen_com} columns 2 and 3, our model consistently outperforms the baseline across on the graph diversity, suggesting that our model successfully learns diverse, real-world strategies for constructing room connection topologies from the training data. 

\begin{figure}[h!]
    \centering
    \includegraphics[width=0.9\textwidth,trim={1cm 0cm 0cm 0cm},clip]{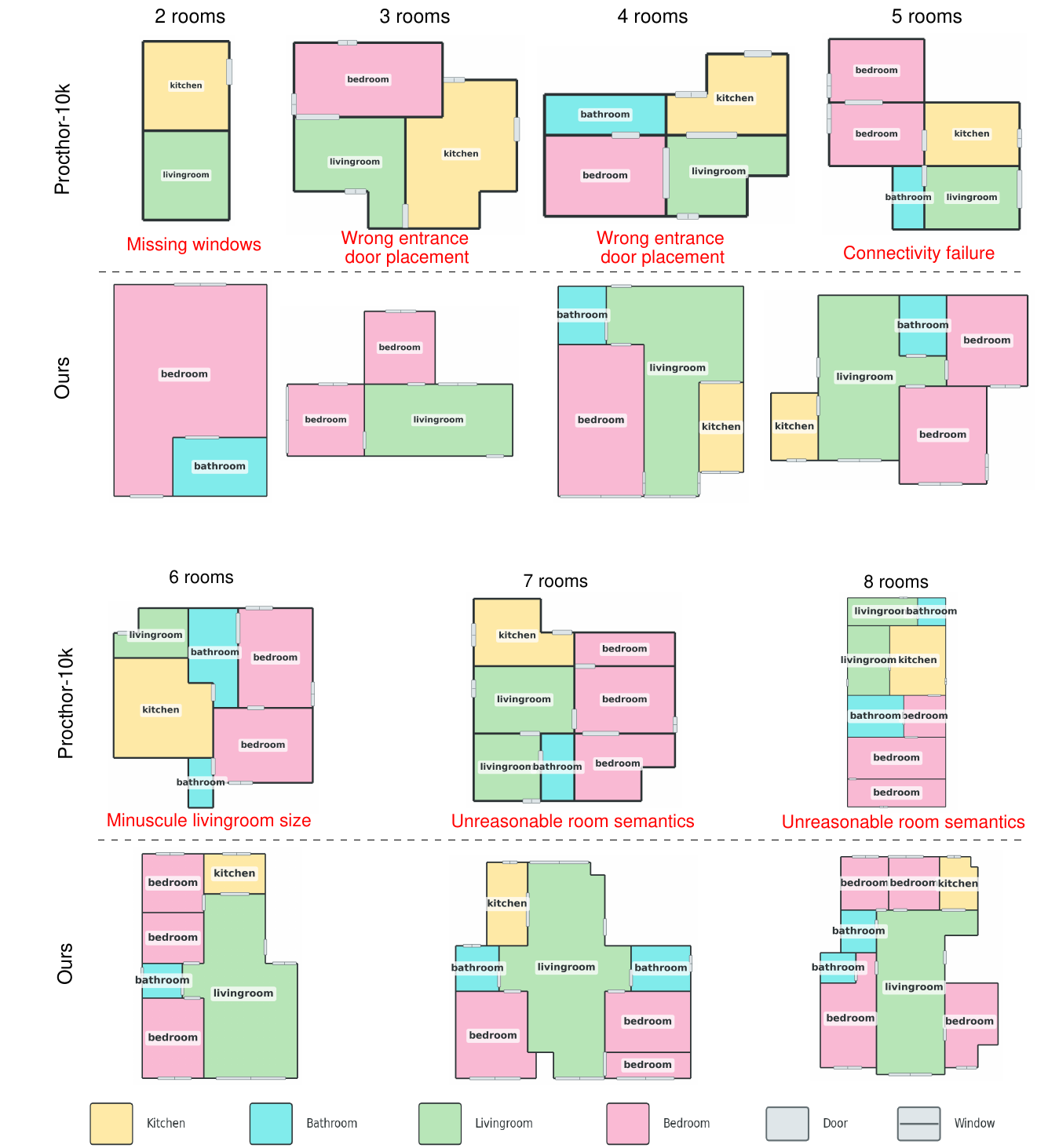}
    \caption{\textbf{Samples from generated floorplan samples. Zoom-in for better view.} We compare our generated floorplans with the baseline across the simpler setting (2-5 rooms) and a more complex setting (6-8 rooms). Our trained model consistently produces more feasible and plausible floorplans compared to that of the baseline method.}
    \label{fig:floorplan_gen}
\end{figure}

\noindent \textbf{Qualitative results.}
We present a qualitative comparison of floorplans generated by our finetuned Qwen3-4B~\cite{qwen3} model against samples from the Procthor~\cite{deitkeProcTHOR️} baseline in~\cref{fig:floorplan_gen}. Examples of structural artifacts include \textit{unreasonable room semnatics}, where doors connect two room types in an implausible manner, such as bedroom–bedroom or livingroom–livingroom connections. \textit{Wrong entrance door placement} refers to positioning the apartment's primary entrance outside the livingroom area. On the other hand, our samples demonstrate superior structural validity and design quality. The baseline method exhibits limited ability to recognize appropriate opening placements, such as \textit{missing window} instances and door placements that disrupt the connectivity topology of the floorplan.

\noindent \textbf{User study.} We conduct a human evaluation study with 24 participants to assess the perceived quality of generated floorplans. We choose the recently proposed FloorPlan-LLaMa~\cite{yin2025floorplan} and Floorplan-Diffusion~\cite{xu2025floorplan} as our compared methods. Participants are presented with three generated floorplans from the three methods and asked to rank them along two dimensions: a) \textit{Reasonability} for practical feasibility, realistic room flows, and constructibility; and b) \textit{Richness} for appropriate organizational richness without unnecessary complication to support multiple daily uses.  To ensure consistent interpretation, all floorplans are rendered using a standardized color palette encoding room types and openings. Rankings are converted to numerical scores: 3, 2, and 1 points for first, second, and third place, respectively. These scores are aggregated across participants and averaged as shown in~\cref{tab:fp_gen_user_study}.
Our generated results consistently achieve the highest quality across both evaluation dimensions.

\begin{table}[htbp]

\caption{
    \textbf{User study of floorplan generation.}
    Our generated results achieve superior quality on both \textit{Reasonability} and \textit{Richness} dimensions.
}
\centering
\scriptsize
\begin{tabular}{lccc}
\toprule
                    & \hspace{5pt}Reasonability\hspace{5pt} & \hspace{5pt}Richness\hspace{5pt} & \hspace{5pt}Average\hspace{5pt} \\
\midrule
FloorPlan-LLaMA~\cite{yin2025floorplan}     & 1.85 & 2.00 & 1.93  \\ 
\midrule
Floorplan-Diffusion~\cite{xu2025floorplan} & 2.13 & 1.81 & 1.97  \\ 
\midrule
Ours                & \textbf{2.20} & \textbf{2.34} & \textbf{2.27} \\
\bottomrule
\end{tabular}

\label{tab:fp_gen_user_study}
\end{table}

\subsection{Furniture Layout Generation}
\begin{figure}[t!]
    \centering

    \includegraphics[width=0.97\textwidth]{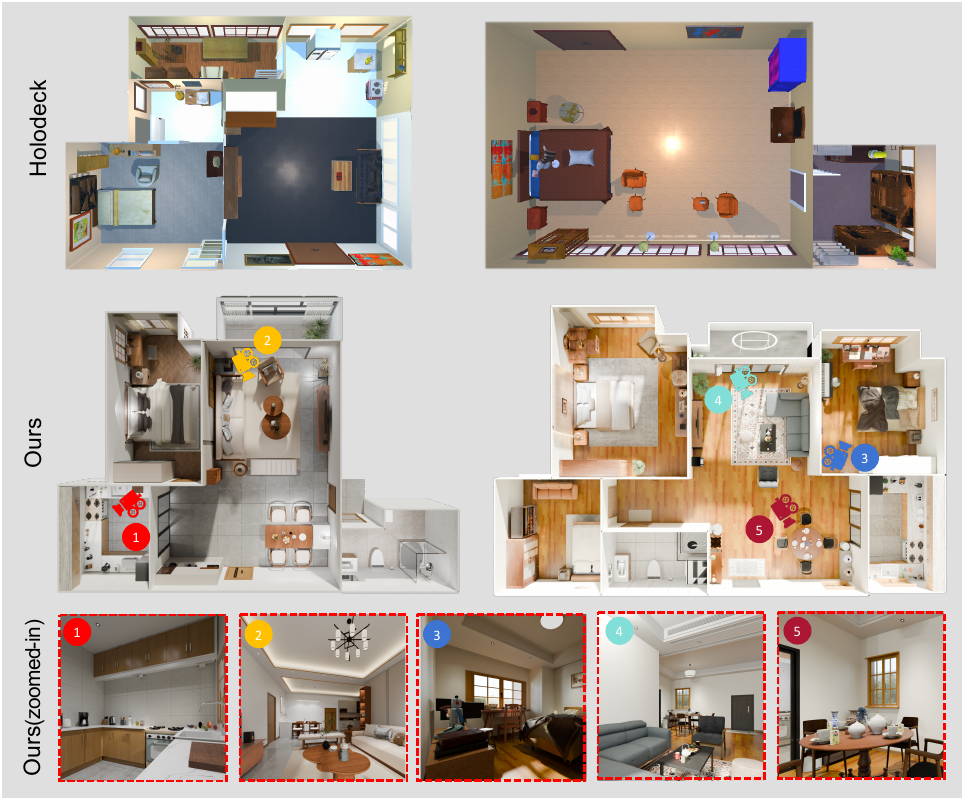}
    \caption{Comparison of whole-home generation.}
    \label{fig:full_flat_2_crop}

\end{figure}
\textbf{Experiment setup.} We exploit Blender to perform all 3D template initialization and perspective rendering in our automated 3D scene generation pipeline.
For 2D image inpainting in both top-down and ego-centric view roaming, we use Gemini 3.0 Image Pro Preview (aka Nano Banana Pro).
For instance-level detection and dense pixel-wise segmentation, we employ SAM3~\cite{carion2025sam}. We then use SAM 3D(Object)~\cite{chen2025sam} to obtain the initial object geometry and pose. 

We evaluate our method against the following baselines: LayoutGPT~\cite{feng2023layoutgpt}, Holodeck~\cite{yang2024holodeck}, and LayoutVLM~\cite{sun2025layoutvlm}, representing existing methods for open-universe layout generation.

\noindent \textbf{Evaluation Metrics.} We evaluate generated 3D layouts along two dimensions: \textbf{physical plausibility} and \textbf{distributional properties}. For physical plausibility, we report the \textit{Collision Ratio} (CR) and the \textit{Out-of-Boundary Ratio} (OOB). CR quantifies severe inter-object penetrations (excluding semantically valid contacts), and OOB measures the fraction of object footprints outside the room boundary. For distributional properties, we report \textit{Volume Density} (VD) and \textit{Footprint Object Density} (FOD), where VD is total 3D box volume normalized by floor area and FOD is the object count normalized by the union area of all furniture footprints, reflecting how densely objects occupy the floor plan.

\begin{figure}[t!]
    \centering
   
    \includegraphics[width=0.9\textwidth]{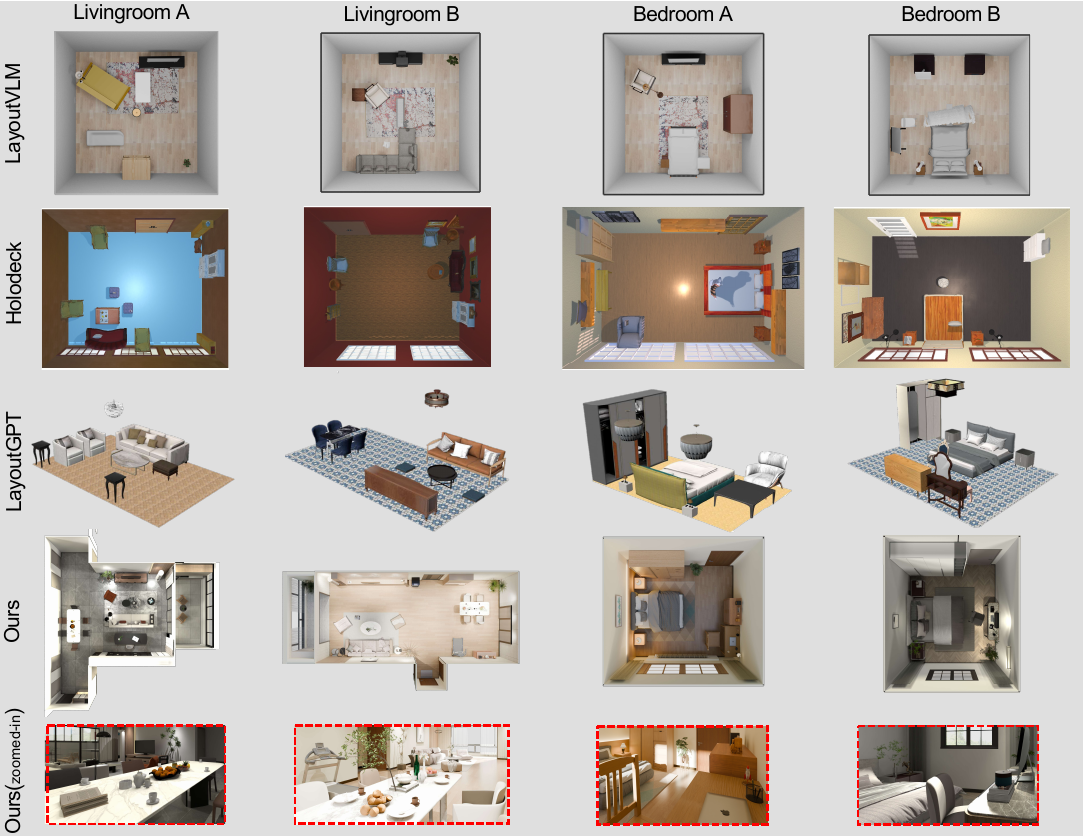}
    \caption{Comparison of generated functional rooms: livingroom and bedroom.}
    \label{fig:livingroom_crop}
\end{figure}

\begin{figure}[h!]
    \centering
    \includegraphics[width=0.9\textwidth]{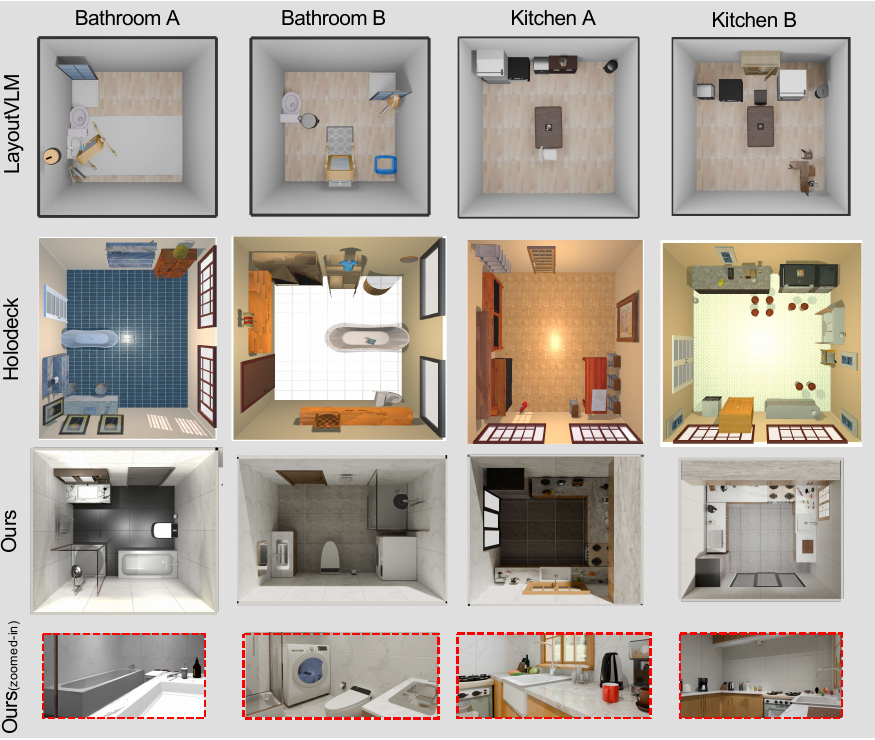}
    \caption{Comparison of generated functional rooms: bathroom and kitchen.}
    \label{fig:bathroom_crop}
\end{figure}

\noindent \textbf{Qualitative results}.We visualize predicted layouts in both whole-home and single functional room settings. Fig.~\ref{fig:full_flat_2_crop} compares whole-home results; among the evaluated baselines, only Holodeck~\cite{yang2024holodeck} and our method support whole-home generation. For single-room generation, we provide additional visual comparisons for the living room/bedroom in Fig.~\ref{fig:livingroom_crop}, and kitchen / bathroom in Fig.~\ref{fig:bathroom_crop}, respectively. Note that LayoutGPT~\cite{feng2023layoutgpt} is limited to living-room and bedroom generation, and thus is excluded from kitchen and bathroom comparisons.

\noindent \textbf{Quantitative results}. To evaluate controllable room-layout generation across common residential spaces, we generate four room categories: living room, bedroom, bathroom, and kitchen, with 25 samples per method per category. 
As shown in Tab.~\ref{tab:layout_generation}, our approach consistently outperforms prior methods on the metrics defined above. 
These quantitative gains align with qualitative observations: our generated layouts exhibit fewer severe overlaps and yield more diverse, natural spatial arrangements.

\begin{table}[htbp]

\caption{\textbf{Quantitative comparison of layout generation.} Our generated results achieve superior quality.}
\centering
\scriptsize
\begin{tabular}{lcccc}
\toprule
& \multicolumn{2}{c}{Physical Plausibility} & \multicolumn{2}{c}{Density \textnormal{\&} Diversity} \\ 
\cmidrule(lr){2-3} \cmidrule(lr){4-5}
Method              & CR$\downarrow$     & OOB$\downarrow$     & VD$\uparrow$   & FOD$\uparrow$    \\ 
\midrule
LayoutGPT~\cite{feng2023layoutgpt}           & 0.05                 & 0.01                 & 0.22           & 1.99    \\
Holodeck~\cite{yang2024holodeck}            & 0.07                 & 0.02                 & 0.22            & 2.10           \\
LayoutVLM~\cite{sun2025layoutvlm}           & 0.20                 & 0.01                 & 0.27            & 2.15           \\
Ours                            &\textbf{0.05}                & \textbf{0.01}                                      & \textbf{0.35}            & \textbf{4.16}  \\
\bottomrule
\end{tabular}

\label{tab:layout_generation}
\end{table}


\noindent \textbf{User Study.} 
Additionally, we conducted a user study participated by 30 users. For each method, we sampled 10 results spanning both the whole-home setting and four single-room categories: living room, bedroom, bathroom, and kitchen.

Participants were shown each sample generated by 4 methods, and asked to rank them from best to worst along three dimensions: (a) \textit{Reasonability} for practical feasibility, realistic room flows, and constructibility; 
(b) \textit{Aesthetics} for visual legibility, balance, and design coherence; 
and (c) \textit{Complexity} for appropriate organizational richness without unnecessary complication.
As reported in Tab.~\ref{tab:layout_user_study}, our method is preferred by roughly all of participants across all questions and scene categories.
\begin{table}[htbp]

\caption{
    \textbf{User study of layout generation. } Our generated results
achieve superior quality on \textit{Reasonability}, \textit{Aesthetics} and \textit{Complexity}.
}
\centering
\scriptsize
\begin{tabular}{lcccc}
\toprule
                    & \hspace{5pt}Reasonability\hspace{5pt} & \hspace{5pt}Aesthetics\hspace{5pt} & \hspace{5pt}Complexity\hspace{5pt} & \hspace{5pt}Average\hspace{5pt} \\
\midrule
LayutGPT~\cite{feng2023layoutgpt}                & 0.202 & 0.184 & 0.153 & 0.180 \\ 
Holodeck~\cite{yang2024holodeck}                & 0.395 & 0.412 & 0.452 & 0.420 \\ 
LayoutVLM~\cite{sun2025layoutvlm}               & 0.252 & 0.260 & 0.283 & 0.265 \\ 
Ours                    & 0.807 & 0.827 & 0.797 & 0.811 \\
\bottomrule
\end{tabular}
\label{tab:layout_user_study}

\end{table}

\subsection{Ablation Study}

\label{sec:ablation}

\noindent \textbf{Ablation on Floorplan Generation}(\textit{w/o K-D tree representation}).
Empirically, when using the canonical representation, 9.2\% of 500 generated floorplans exhibit overlapping room polygons, and 2.0\% contain holes within apartment outlines. Furthermore, the percentage of floorplans with disconnected rooms rises to 16.2\% under the canonical representation, compared to 12.3\% with our K-D tree formulation-a notable degradation in the reliability and structural coherence of floorplan generation.

\begin{table}[htbp]

\caption{\textbf{Ablation study on the Layout Generation.}}
\centering
\scriptsize
\begin{tabular}{lcccc}
\toprule
                        & \multicolumn{2}{c}{Physical Plausibility} & \multicolumn{2}{c}{Density \textnormal{\&} Diversity} \\ 
                        \cmidrule(lr){2-3}                              \cmidrule(lr){4-5}
Method                  & CR$\downarrow$     & OOB$\downarrow$      & VD$\uparrow$                  & FOD$\uparrow$    \\ 
\midrule
w/o Top-down View Roaming      & 0.09                & 0.02                  & 0.22             & 2.6           \\
w/o Ego-centric View Roaming    & 0.07                & 0.02                  & 0.20             & 3.44           \\
w/o VLM–LLM Refiner             & 0.20                & 0.05                  & 0.33             & 3.02           \\
w/o Manipulable Object Placement    & 0.05                & 0.01                  & 0.33             & 1.82           \\
\textbf{Ours}           & 0.05                & 0.01                  & 0.35             & 4.16           \\
\bottomrule
\end{tabular}

\label{tab:layout_ablation}
\end{table}

\noindent \textbf{Ablation on Layout Generation.} We conduct an ablation study by disabling one module at a time while keeping all remaining components, prompts, and the evaluation protocol unchanged. We compare four variants against the full system (\textbf{Ours}) in Tab.~\ref{tab:layout_ablation}, reporting physical plausibility (\textbf{CR}$\downarrow$, \textbf{OOB}$\downarrow$) and distributional properties (\textbf{VD}$\uparrow$, \textbf{FOD}$\uparrow$).

1)\textit{ w/o top-down view roaming.} We remove the structure-aware top-down stage (top-view inpainting and the top-down 2D-to-3D initialization) and rely on the remaining pipeline. This leads to worse reduces scene density (\textbf{VD} drops from \textbf{0.35} to \textbf{0.22}), indicating that the top-down stage provides crucial global structure priors for both feasibility and coarse furniture placement.

2)\textit{ w/o ego-centric view roaming.} We disable ego-centric roaming and generate layouts only from the top-down initialization. The plausibility slightly degrades, while \textbf{VD} decreases (\textbf{0.35} $\rightarrow$ \textbf{0.20}), showing that ego-centric roaming is important for enriching furniture-level content.

3)\textit{ w/o VLM refiner.} We turn off the recursive refinement loop (no multi-view consistency checking or iterative box corrections). This significantly harms physical plausibility (\textbf{CR/OOB} rises to \textbf{0.20/0.05}), demonstrating that the refiner is critical for correcting structural and physical violations produced by synthesis and grounding.

4)\textit{ w/o manipulable object placement.} We remove manipulable surface-object placement while keeping the rest of the pipeline unchanged. Physical plausibility remains on par with \textbf{Ours}, but \textbf{FOD} drops sharply (\textbf{4.16} $\rightarrow$ \textbf{1.82}) and \textbf{VD} slightly decreases (\textbf{0.35} $\rightarrow$ \textbf{0.33}), indicating that object placement mainly increases the number of small, manipulable objects without sacrificing feasibility.

Overall, the top-down stage and ego-centric stage improves global feasibility and coarse density, object placement substantially increase object-level richness (especially FOD), and the VLM-based refiner is essential for ensuring physical plausibility.

\section{Conclusion}
In this work, we present HomeWorld, a unified hierarchical framework for controllable whole-home scene generation from natural-language prompts. First, we synthesize fine-grained residential floorplans using an LLM trained on a large-scale real-world floorplan corpus with a structured K-D tree representation. We then progressively instantiate the generated layout into a furnished 3D home through hierarchical view roaming, recursive refinement, and surface-centric manipulable object placement. This unified design connects floorplan generation with downstream 3D scene realization, enabling the creation of complete whole-home environments rather than isolated rooms.

To address the scarcity of large-scale 3D residential training data, we combine scalable 2D generative priors with explicit 3D grounding and iterative correction. The resulting pipeline supports coherent furniture arrangement, realistic placement of manipulable small objects, and basic scene completion with physical attributes, texture assignment, and lighting setup for downstream embodied AI simulation. Experiments and user studies show that our framework produces more valid and diverse floorplans, as well as more plausible and richer 3D layouts, than prior methods.

Alongside the generation framework, we will release a dataset of 300K annotated real-world residential floorplans together with 5K high-quality furnished whole-home scenes. We hope this combination of model and data can support future research on controllable indoor scene generation and simulation-ready embodied AI environments.

\newpage
\bibliographystyle{unsrt}
\bibliography{main}

@String(ICLR  = {Int. Conf. Learn. Represent.})

@String(TOG   = {ACM Trans. Graph.})

@String(ICLR  = {ICLR})

@String(TOG   = {ACM TOG})

@article{chang2017matterport3d,
  title={Matterport3d: Learning from rgb-d data in indoor environments},
  author={Chang, Angel and Dai, Angela and Funkhouser, Thomas and Halber, Maciej and Niessner, Matthias and Savva, Manolis and Song, Shuran and Zeng, Andy and Zhang, Yinda},
  journal={arXiv preprint arXiv:1709.06158},
  year={2017}
}

@inproceedings{fu20213d,
  title={3d-front: 3d furnished rooms with layouts and semantics},
  author={Fu, Huan and Cai, Bowen and Gao, Lin and Zhang, Ling-Xiao and Wang, Jiaming and Li, Cao and Zeng, Qixun and Sun, Chengyue and Jia, Rongfei and Zhao, Binqiang and others},
  booktitle={Proceedings of the IEEE/CVF International Conference on Computer Vision},
  pages={10933--10942},
  year={2021}
}

@inproceedings{dai2017scannet,
  title={Scannet: Richly-annotated 3d reconstructions of indoor scenes},
  author={Dai, Angela and Chang, Angel X and Savva, Manolis and Halber, Maciej and Funkhouser, Thomas and Nie{\ss}ner, Matthias},
  booktitle={Proceedings of the IEEE conference on computer vision and pattern recognition},
  pages={5828--5839},
  year={2017}
}

@inproceedings{zheng2020structured3d,
  title={Structured3d: A large photo-realistic dataset for structured 3d modeling},
  author={Zheng, Jia and Zhang, Junfei and Li, Jing and Tang, Rui and Gao, Shenghua and Zhou, Zihan},
  booktitle={European Conference on Computer Vision},
  pages={519--535},
  year={2020},
  organization={Springer}
}

@article{wu2024blockfusion,
  title={Blockfusion: Expandable 3d scene generation using latent tri-plane extrapolation},
  author={Wu, Zhennan and Li, Yang and Yan, Han and Shang, Taizhang and Sun, Weixuan and Wang, Senbo and Cui, Ruikai and Liu, Weizhe and Sato, Hiroyuki and Li, Hongdong and others},
  journal={ACM Transactions on Graphics (ToG)},
  volume={43},
  number={4},
  pages={1--17},
  year={2024},
  publisher={ACM New York, NY, USA}
}

@inproceedings{ju2024diffindscene,
  title={Diffindscene: Diffusion-based high-quality 3d indoor scene generation},
  author={Ju, Xiaoliang and Huang, Zhaoyang and Li, Yijin and Zhang, Guofeng and Qiao, Yu and Li, Hongsheng},
  booktitle={Proceedings of the IEEE/CVF Conference on Computer Vision and Pattern Recognition},
  pages={4526--4535},
  year={2024}
}

@article{deitkeProcTHOR️,
  title={ProcTHOR: Large-Scale Embodied AI Using Procedural Generation},
  author={Deitke, Matt and VanderBilt, Eli and Herrasti, Alvaro and Weihs, Luca and Ehsani, Kiana and Salvador, Jordi and Han, Winson and Kolve, Eric and Kembhavi, Aniruddha and Mottaghi, Roozbeh},
  journal={Advances in Neural Information Processing Systems},
  volume={35},
  pages={5982--5994},
  year={2022}
}

@article{savva2017minos,
  title={MINOS: Multimodal indoor simulator for navigation in complex environments},
  author={Savva, Manolis and Chang, Angel X and Dosovitskiy, Alexey and Funkhouser, Thomas and Koltun, Vladlen},
  journal={arXiv preprint arXiv:1712.03931},
  year={2017}
}

@inproceedings{hollein2023text2room,
  title={Text2room: Extracting textured 3d meshes from 2d text-to-image models},
  author={H{\"o}llein, Lukas and Cao, Ang and Owens, Andrew and Johnson, Justin and Nie{\ss}ner, Matthias},
  booktitle={Proceedings of the IEEE/CVF International Conference on Computer Vision},
  pages={7909--7920},
  year={2023}
}

@misc{LIFULLNII2017LIFULLHRFP,
  author       = {{LIFULL Co., Ltd.} and {National Institute of Informatics (NII)}},
  title        = {{LIFULL HOME'S High Resolution Floor Plan Image Data}},
  year         = {2017},
  howpublished = {\url{https://www.nii.ac.jp/dsc/idr/en/lifull/1.html}},
  note         = {Accessed: 2026-02-28}
}

@article {Wu_DeepLayout_2019,
title = {Data-driven Interior Plan Generation for Residential Buildings},
author = {Wenming Wu and Xiao-Ming Fu and Rui Tang and Yuhan Wang and Yu-Hao Qi and Ligang Liu},
journal = {ACM Transactions on Graphics (SIGGRAPH Asia)},
volume = {38},
number = {6},
year = {2019},
}

@inproceedings{van2024msd,
  title={Msd: A benchmark dataset for floor plan generation of building complexes},
  author={Van Engelenburg, Casper and Mostafavi, Fatemeh and Kuhn, Emanuel and Jeon, Yuntae and Franzen, Michael and Standfest, Matthias and van Gemert, Jan and Khademi, Seyran},
  booktitle={European Conference on Computer Vision},
  pages={60--75},
  year={2024},
  organization={Springer}
}

@article{abouagour2025resplan,
  title={ResPlan: A Large-Scale Vector-Graph Dataset of 17,000 Residential Floor Plans},
  author={Abouagour, Mohamed and Garyfallidis, Eleftherios},
  journal={arXiv preprint arXiv:2508.14006},
  year={2025}
}

@article{zhong2025internscenes,
  title={Internscenes: A large-scale simulatable indoor scene dataset with realistic layouts},
  author={Zhong, Weipeng and Cao, Peizhou and Jin, Yichen and Luo, Li and Cai, Wenzhe and Lin, Jingli and Wang, Hanqing and Lyu, Zhaoyang and Wang, Tai and Dai, Bo and others},
  journal={arXiv preprint arXiv:2509.10813},
  year={2025}
}

@inproceedings{jia2024sceneverse,
  title={Sceneverse: Scaling 3d vision-language learning for grounded scene understanding},
  author={Jia, Baoxiong and Chen, Yixin and Yu, Huangyue and Wang, Yan and Niu, Xuesong and Liu, Tengyu and Li, Qing and Huang, Siyuan},
  booktitle={European Conference on Computer Vision},
  pages={289--310},
  year={2024},
  organization={Springer}
}

@inproceedings{raistrick2024infinigen,
  title={Infinigen indoors: Photorealistic indoor scenes using procedural generation},
  author={Raistrick, Alexander and Mei, Lingjie and Kayan, Karhan and Yan, David and Zuo, Yiming and Han, Beining and Wen, Hongyu and Parakh, Meenal and Alexandropoulos, Stamatis and Lipson, Lahav and others},
  booktitle={Proceedings of the IEEE/CVF Conference on Computer Vision and Pattern Recognition},
  pages={21783--21794},
  year={2024}
}

@inproceedings{bokhovkin2025scenefactor,
  title={Scenefactor: Factored latent 3d diffusion for controllable 3d scene generation},
  author={Bokhovkin, Aleksey and Meng, Quan and Tulsiani, Shubham and Dai, Angela},
  booktitle={Proceedings of the Computer Vision and Pattern Recognition Conference},
  pages={628--639},
  year={2025}
}

@inproceedings{tang2024diffuscene,
  title={Diffuscene: Denoising diffusion models for generative indoor scene synthesis},
  author={Tang, Jiapeng and Nie, Yinyu and Markhasin, Lev and Dai, Angela and Thies, Justus and Nie{\ss}ner, Matthias},
  booktitle={Proceedings of the IEEE/CVF conference on computer vision and pattern recognition},
  pages={20507--20518},
  year={2024}
}

@article{chung2023luciddreamer,
  title={Luciddreamer: Domain-free generation of 3d gaussian splatting scenes},
  author={Chung, Jaeyoung and Lee, Suyoung and Nam, Hyeongjin and Lee, Jaerin and Lee, Kyoung Mu},
  journal={arXiv preprint arXiv:2311.13384},
  year={2023}
}

@inproceedings{chen2025housecrafter,
  title={HouseCrafter: Lifting Floorplans to 3D Scenes with 2D Diffusion Models},
  author={Chen, Yiwen and Nguyen, Hieu T and Voleti, Vikram and Jampani, Varun and Jiang, Huaizu},
  booktitle={Proceedings of the IEEE/CVF International Conference on Computer Vision},
  pages={28440--28450},
  year={2025}
}

@article{feng2023layoutgpt,
  title={Layoutgpt: Compositional visual planning and generation with large language models},
  author={Feng, Weixi and Zhu, Wanrong and Fu, Tsu-jui and Jampani, Varun and Akula, Arjun and He, Xuehai and Basu, Sugato and Wang, Xin Eric and Wang, William Yang},
  journal={Advances in Neural Information Processing Systems},
  volume={36},
  pages={18225--18250},
  year={2023}
}

@inproceedings{sun2025layoutvlm,
  title={Layoutvlm: Differentiable optimization of 3d layout via vision-language models},
  author={Sun, Fan-Yun and Liu, Weiyu and Gu, Siyi and Lim, Dylan and Bhat, Goutam and Tombari, Federico and Li, Manling and Haber, Nick and Wu, Jiajun},
  booktitle={Proceedings of the Computer Vision and Pattern Recognition Conference},
  pages={29469--29478},
  year={2025}
}

@inproceedings{yang2024holodeck,
  title={Holodeck: Language guided generation of 3d embodied ai environments},
  author={Yang, Yue and Sun, Fan-Yun and Weihs, Luca and VanderBilt, Eli and Herrasti, Alvaro and Han, Winson and Wu, Jiajun and Haber, Nick and Krishna, Ranjay and Liu, Lingjie and others},
  booktitle={Proceedings of the IEEE/CVF Conference on Computer Vision and Pattern Recognition},
  pages={16227--16237},
  year={2024}
}

@inproceedings{ccelen2024design,
  title={I-design: Personalized llm interior designer},
  author={{\c{C}}elen, Ata and Han, Guo and Schindler, Konrad and Van Gool, Luc and Armeni, Iro and Obukhov, Anton and Wang, Xi},
  booktitle={European Conference on Computer Vision},
  pages={217--234},
  year={2024},
  organization={Springer}
}

@article{liu2025worldcraft,
  title={Worldcraft: Photo-realistic 3d world creation and customization via llm agents},
  author={Liu, Xinhang and Tang, Chi-Keung and Tai, Yu-Wing},
  journal={arXiv preprint arXiv:2502.15601},
  year={2025}
}

@article{yang2024llplace,
  title={Llplace: The 3d indoor scene layout generation and editing via large language model},
  author={Yang, Yixuan and Lu, Junru and Zhao, Zixiang and Luo, Zhen and Yu, James JQ and Sanchez, Victor and Zheng, Feng},
  journal={arXiv preprint arXiv:2406.03866},
  year={2024}
}

@inproceedings{huang2025midi,
  title={Midi: Multi-instance diffusion for single image to 3d scene generation},
  author={Huang, Zehuan and Guo, Yuan-Chen and An, Xingqiao and Yang, Yunhan and Li, Yangguang and Zou, Zi-Xin and Liang, Ding and Liu, Xihui and Cao, Yan-Pei and Sheng, Lu},
  booktitle={Proceedings of the IEEE/CVF Conference on Computer Vision and Pattern Recognition},
  pages={23646--23657},
  year={2025}
}

@article{bucher2025respace,
  title={ReSpace: Text-Driven 3D Indoor Scene Synthesis and Editing with Preference Alignment},
  author={Bucher, Martin JJ and Armeni, Iro},
  journal={arXiv preprint arXiv:2506.02459},
  year={2025}
}

@article{carion2025sam,
  title={Sam 3: Segment anything with concepts},
  author={Carion, Nicolas and Gustafson, Laura and Hu, Yuan-Ting and Debnath, Shoubhik and Hu, Ronghang and Suris, Didac and Ryali, Chaitanya and Alwala, Kalyan Vasudev and Khedr, Haitham and Huang, Andrew and others},
  journal={arXiv preprint arXiv:2511.16719},
  year={2025}
}

@article{chen2025sam,
  title={Sam 3d: 3dfy anything in images},
  author={Chen, Xingyu and Chu, Fu-Jen and Gleize, Pierre and Liang, Kevin J and Sax, Alexander and Tang, Hao and Wang, Weiyao and Guo, Michelle and Hardin, Thibaut and Li, Xiang and others},
  journal={arXiv preprint arXiv:2511.16624},
  year={2025}
}

@inproceedings{yin2025floorplan,
  title={Floorplan-llama: Aligning architects’ feedback and domain knowledge in architectural floor plan generation},
  author={Yin, Jun and Zeng, Pengyu and Sun, Haoyuan and Dai, Yuqin and Zheng, Han and Zhang, Miao and Zhang, Yachao and Lu, Shuai},
  booktitle={Proceedings of the 63rd Annual Meeting of the Association for Computational Linguistics (Volume 1: Long Papers)},
  pages={6640--6662},
  year={2025}
}

@inproceedings{xu2025floorplan,
  title={Floorplan-Diffusion: Automatic Floor Plan Generation via Pre-trained Large Latent Diffusion Model},
  author={Xu, Minyang and Lou, Yunzhong and Gao, Xiang and Zhou, Xiangdong},
  booktitle={Proceedings of the 2025 International Conference on Multimedia Retrieval},
  pages={1617--1625},
  year={2025}
}

@article{qwen3,
    title={Qwen3 Technical Report}, 
    author={An Yang and Anfeng Li and Baosong Yang and Beichen Zhang and Binyuan Hui and Bo Zheng and Bowen Yu and Chang Gao and Chengen Huang and Chenxu Lv and Chujie Zheng and Dayiheng Liu and Fan Zhou and Fei Huang and Feng Hu and Hao Ge and Haoran Wei and Huan Lin and Jialong Tang and Jian Yang and Jianhong Tu and Jianwei Zhang and Jianxin Yang and Jiaxi Yang and Jing Zhou and Jingren Zhou and Junyang Lin and Kai Dang and Keqin Bao and Kexin Yang and Le Yu and Lianghao Deng and Mei Li and Mingfeng Xue and Mingze Li and Pei Zhang and Peng Wang and Qin Zhu and Rui Men and Ruize Gao and Shixuan Liu and Shuang Luo and Tianhao Li and Tianyi Tang and Wenbiao Yin and Xingzhang Ren and Xinyu Wang and Xinyu Zhang and Xuancheng Ren and Yang Fan and Yang Su and Yichang Zhang and Yinger Zhang and Yu Wan and Yuqiong Liu and Zekun Wang and Zeyu Cui and Zhenru Zhang and Zhipeng Zhou and Zihan Qiu},
    journal = {arXiv preprint arXiv:2505.09388},
    year={2025}
}

@article{hu2022lora,
  title={Lora: Low-rank adaptation of large language models.},
  author={Hu, Edward J and Shen, Yelong and Wallis, Phillip and Allen-Zhu, Zeyuan and Li, Yuanzhi and Wang, Shean and Wang, Liang and Chen, Weizhu and others},
  journal={Iclr},
  volume={1},
  number={2},
  pages={3},
  year={2022}
}

@inproceedings{yang2024physcene,
  title={Physcene: Physically interactable 3d scene synthesis for embodied ai},
  author={Yang, Yandan and Jia, Baoxiong and Zhi, Peiyuan and Huang, Siyuan},
  booktitle={Proceedings of the IEEE/CVF Conference on Computer Vision and Pattern Recognition},
  pages={16262--16272},
  year={2024}
}

@article{su2025chord,
  title={Chord: Generation of collision-free, house-scale, and organized digital twins for 3d indoor scenes with controllable floor plans and optimal layouts},
  author={Su, Chong and Fu, Yingbin and Hu, Zheyuan and Yang, Jing and Hanji, Param and Wang, Shaojun and Zhao, Xuan and {\"O}ztireli, Cengiz and Zhong, Fangcheng},
  journal={arXiv preprint arXiv:2503.11958},
  year={2025}
}

@article{wang2025embodiedgen,
  title={Embodiedgen: Towards a generative 3d world engine for embodied intelligence},
  author={Wang, Xinjie and Liu, Liu and Cao, Yu and Wu, Ruiqi and Qin, Wenkang and Wang, Dehui and Sui, Wei and Su, Zhizhong},
  journal={arXiv preprint arXiv:2506.10600},
  year={2025}
}

@article{cao2025physx,
  title={PhysX-Anything: Simulation-Ready Physical 3D Assets from Single Image},
  author={Cao, Ziang and Hong, Fangzhou and Chen, Zhaoxi and Pan, Liang and Liu, Ziwei},
  journal={arXiv preprint arXiv:2511.13648},
  year={2025}
}

\end{document}